\documentclass[sigconf]{acmart}

\usepackage{booktabs} 
\usepackage{multirow}
\usepackage{tabularx}

\setcopyright{rightsretained}

\acmDOI{}

\acmISBN{}

\acmConference[]{}{}{}
\acmYear{}
\copyrightyear{}

\begin{document}
\settopmatter{printacmref=false} 
\renewcommand\footnotetextcopyrightpermission[1]{} 
\pagestyle{plain} 
\title{An Attention Model for group-level emotion recognition }

\thanks{*These authors contributed equally}

\author{Aarush Gupta*}
\affiliation{%
  \institution{Indian Institute of Technology Roorkee}
  \city{Roorkee}
  \country{India}
}
\email{agupta1@cs.iitr.ac.in}
\author{Dakshit Agrawal*}
\affiliation{%
  \institution{Indian Institute of Technology Roorkee}
  \city{Roorkee}
  \country{India}
} 
\email{dagrawal@cs.iitr.ac.in}

\author{Hardik Chauhan}
\affiliation{%
  \institution{Indian Institute of Technology Roorkee}
  \city{Roorkee}
  \country{India}}
\email{haroi.uee2014@iitr.ac.in}

\author{Jose Dolz}
\affiliation{%
  \institution{\'Ecole de Technologie Sup\'erieure}
  \city{Montreal}
  \country{Canada}}
\email{jose.dolz@livia.etsmtl.ca}

\author{Marco Pedersoli}
\affiliation{%
  \institution{\'Ecole de Technologie Sup\'erieure}
  \city{Montreal}
  \country{Canada}}
\email{Marco.Pedersoli@etsmtl.ca}

\renewcommand{\shortauthors}{Anonymous}

\begin{abstract}
In this paper we propose a new approach for classifying the global emotion of images containing groups of people. To achieve this task, we consider two different and complementary sources of information: i) a global representation of the entire image (ii) a local representation where only faces are considered. While the global representation of the image is learned with a convolutional neural network (CNN), the local representation is obtained by merging face features through an attention mechanism. 
The two representations are first learned independently with two separate CNN branches and then fused through concatenation in order to obtain the final group-emotion classifier.
For our submission to the EmotiW 2018 group-level emotion recognition challenge, we combine several variations of the proposed model into an ensemble, obtaining a final accuracy of 64.83\% on the test set and ranking 4th among all challenge participants. 

\end{abstract}

%
%



\maketitle

\section{Introduction}
The recognition of emotions is still a challenging task, despite the interest shown in this problem. Researchers obtained promising results investigating some very specific features, such as heartbeat and blood pressure ~\cite{predicting15Jaques,Choi2017IsHR}. However, an important limitation is that these features can only be obtained with specific wearable devices and therefore in very controlled settings. Recently, modern computer vision approaches \cite{EbrahimiKahou15RNN} have achieved outstanding performance in this task from passive sensors signals, such as audio-visual data, opening the door to emotion recognition in uncontrolled settings, i.e. in the wild. Despite these advances, for good performance, it is still necessary to provide the method with clean audio and a close-up of the human face, so that it can extract salient features about the person emotions {\cite{DBLP:journals/corr/KaraliBE16}.

In this work, we consider an even more challenging task of classifying the emotion shared among a group of people in an image as either positive, neutral or negative. This task is commonly referred to as group-level emotion recognition \cite{DBLP:EmotiW2018}. 
Compared to the recognition of emotions in videos, the group-level emotion} recognition is more challenging because it is based on a single image (lack of temporal information) and the human faces are often at low-resolution (lack of facial details). Nonetheless, the emotion is shared across a group of people, and the environment can also help to recognize the correct emotion to some extent. 

Previous work on this task has focused on taking the scene and facial features of an image \cite{Tan:2017:GER:3136755.3143008, Wei:2017:NDF:3136755.3143014}, as well as the pose of people and/or their faces \cite{Guo:2017:GER:3136755.3143017}. These approaches, however, do not learn how to appropriately combine the information coming from the different faces and the global image. 
A natural way of doing that is by employing a mechanism of attention \cite{Bahdanau2014NeuralMT} that is capable of ranking faces importance. Inspired by the selective attention of human perception, different mechanisms of attention have been proposed for computer vision problems -- e.g. image classification \cite{mnih2014recurrent,wang2017residual}, action recognition \cite{girdhar2017attentional}, and image caption generation \cite{xu2015show}-- with the goal of selecting the relevant regions of an image or a video and substantially reducing the task complexity. 




In this paper, we present an attention mechanism to combine the local representation of all faces. To achieve this, we build an auxiliary network that learns to associate an importance score with each face in the image. This score is then used to combine the facial features into a single representation. We also propose an end-to-end model for learning the scene and facial features of an image jointly. In the experimental results, we show that such representation is better for group-emotion recognition than averaging \cite{Tan:2017:GER:3136755.3143008}. 
 Finally, to reduce over-fitting, we pre-train our networks on a larger dataset similar to the EmotiW 2018 challenge dataset \cite{7163151} and combine several models in an ensemble. This corresponds to our EmotiW 2018 challenge submission that obtained an accuracy of 64.83\% on the test set and ranked 4th among all challenge participants.

\begin{figure*}
\includegraphics{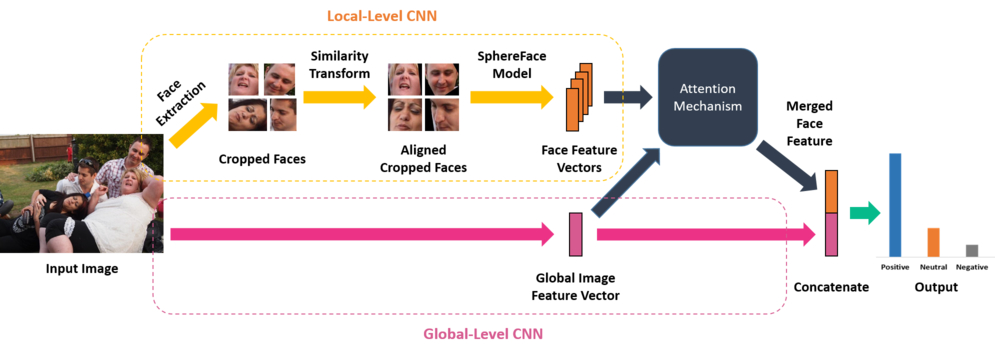}
\caption{Overview of proposed Attention Model, a two-branched CNN model that learns the global and local features jointly.}
\label{fig:modeloverview}
\end{figure*}

\section{Materials and Methods}




\subsection{Our Approach}

We propose an end-to-end model for jointly learning the scene and facial features of an image  for group-level emotion recognition. An overview of the approach is presented in Fig.~\ref{fig:modeloverview}. Our model is composed of two branches.  The first branch is a global-level CNN (sec.~\ref{sec:global}) that detects emotions on the basis of the image as a whole.  The second is a local-level CNN (sec.~\ref{sec:local}) that detects emotions on the basis of the faces present in the image. The content of each face is merged into a single representation by an attention mechanism (sec.~\ref{sec:attention}).  This single representation of the facial features is then concatenated with the image feature vector from the Global-Level CNN to build an end-to-end trainable model. In the following subsections, we briefly describe each part of our model.

%
%
%

\subsection{Global-Level CNN}
\label{sec:global}
The surroundings in which the group photo is taken can be important for recognizing the emotion that is portrayed by the group.  For example, a photo taken during a funeral is most likely to depict a negative emotion.  Similarly, a photo taken in a marriage is most likely to show a positive emotion.  Motivated by this, we employ a state-of-the-art classification network, i.e., DenseNet-161 \cite{DBLP:journals/corr/HuangLW16a}, to learn global features from whole images. The network is pre-trained on ImageNet \cite{imagenet_cvpr09}.

\subsection{Local-Level CNN}
\label{sec:local}
In addition to global context, the emotions portrayed by faces in a group image play a crucial role in emotion recognition.  Hence, we also train deep learning models that predict the emotion of the image based on the faces present in that image.  Thus, inspired by last year's challenge winners \cite{Tan:2017:GER:3136755.3143008}, we build a local-level CNN that analyses the emotion of each individual's face in the image.

\subsubsection{Face Extraction and Alignment}

We use the Multi-Task Cascaded Convolutional Network model (MTCNN) \cite{DBLP:journals/corr/ZhangZL016} to extract faces from the image because of its high performance and speed. MTCNN detects the face bounding boxes as well as the corresponding facial landmarks.

The faces obtained from the MTCNN model have different orientations and scales according to the given image.  
Due to the small size of the challenge dataset, in order to learn a simpler model, we normalize each face to the frontal view and represent the facial image with a fixed resolution. In practice, we apply a similarity transform using the facial landmarks such that the eyes of the faces are at the horizontal level and the image is rescaled to the size of 96$\times$112.

\subsubsection{CNN Model for Face Emotion Recognition}

The aligned faces are then passed through SphereFace \cite{DBLP:journals/corr/LiuWYLRS17}, a CNN model pre-trained on the CASIA-Webface dataset \cite{DBLP:journals/corr/YiLLL14a}. We use the model with pre-trained weights as it usually gives better performance.

\subsection{Attention Mechanisms}
\label{sec:attention}
From the local-level CNN, we obtain a different representation for each face. However, we need to convert it into a single representation that can be evaluated independently of the number of faces that are present in the image. In this sense, a simple concatenation of the feature descriptors will not work well, because each image can contain a different number of faces and this is unknown apriori.

The simplest solution is to compute the average features as in \cite{Tan:2017:GER:3136755.3143008}.
However, not all faces are equally important for recognizing the emotion of the image. Some faces have more significance in portraying the actual emotion, while others can confuse the final classification.  
For instance, consider the case of a cry-laugh. Many methods may easily confuse it as a negative emotion, hence failing the prediction. In contrast, if we associate a confidence value with each face in the image, we can still infer that the image represents a positive emotion, by assigning a low importance to the crying face.
Based on this observation, we use attention mechanisms \cite{Bahdanau2014NeuralMT, DBLP:journals/corr/VaswaniSPUJGKP17} to find probabilistic weights for each face in the image.  A weighted sum according to these weights is computed to produce a single representation of the facial features.

\begin{figure*}
\includegraphics{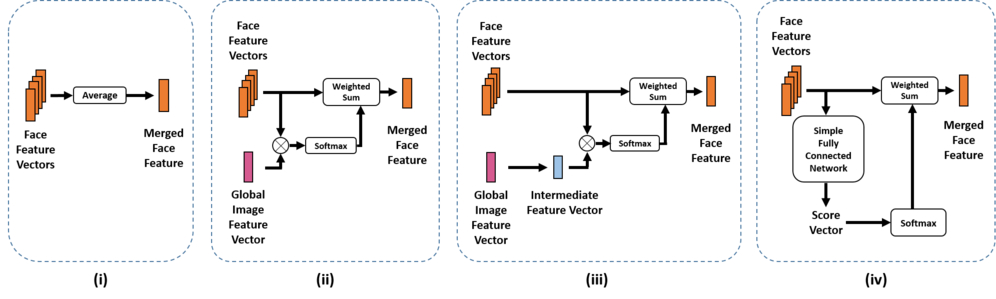}
\caption{Attention Mechanisms to merge the Face Feature Vectors:  (i) Average of Facial Feature Vectors  (ii) Attention A: Global Image Feature Vector  (iii) Attention B: Intermediate Feature Vector  (iv) Attention C: Feature Score}
\label{fig:attentionmechanism}
\end{figure*}

As shown in Fig.~\ref{fig:attentionmechanism}, in our experiments, we consider four different ways to merge the face feature vector of each individual in the image.

\subsubsection{Average Features}\label{sec:mechansism1}

As shown in Fig.~2(i) and similar to \cite{Tan:2017:GER:3136755.3143008}, we simply compute the average of the face feature vectors from the local-level CNN to obtain a single facial feature vector.

\subsubsection{Attention A: Global Image Feature Vector}\label{sec:mechanism2}

The scheme for this attention mechanism is shown in Fig.~2(ii).  The image feature vector obtained from the Global-Level CNN is used as the query vector and the face features obtained from the Local-Level CNN are used as the key vectors (one face feature is one key vector).  The dot product between the face feature vectors and the global image feature vector is computed followed by softmax to produce the attention weight for each face. The weighted sum of the face features is obtained according to the attention weights to produce a single facial feature vector representation.
This attention mechanism assigns more weight to the face feature vector that has higher similarity to the global image feature vector.

\subsubsection{Attention B: Intermediate Feature Vector}\label{sec:mechanism3}

The scheme for this attention mechanism is shown in Fig.~ 2(iii).  It is similar to Attention A (see sec.~ \ref{sec:mechanism2}).  The difference is that the global-level CNN representation is transformed by an intermediate fully connected layer into a more compact representation. The intermediate feature vector layer converts the global image feature vector into a representation that is more suitable for comparing facial features.

\subsubsection{Attention C: Feature Score}\label{sec:mechanism4}

The scheme for this attention mechanism is shown in Fig.~ 2(iv).  The face feature vectors are passed through a simple fully connected network to obtain a score vector, followed by Softmax function to produce the probabilistic attention weights.  A weighted sum of the face features is computed according to the attention weights to produce a single facial feature vector representation. While in the previous models the attention was generated based on a global representation of the image, in this case, the attention is learned by the fully connected neural network.

\section{Experiments and Results}

\subsection{Dataset}
The task of the EmotiW 2018 group-level emotion recognition challenge is to classify a group's perceived emotion as \textit{Positive, Neutral or Negative}.  The images in this sub-challenge are from the Group Affect Database 2.0 \cite{7163151}. It consists of 9,815 train images, 4,346 validation images and 3,011 test images.  Note that the annotations for test images are not available.


We further split the validation data into two parts. The first part (VAL), composed of 3,346 images, is used for tuning the hyper-parameters of our models, while the second part (EVAL), composed of 1,000 images, is used for the final evaluation of the different models.

\subsection{Experimental Results}

\subsubsection{Global-Level CNN} 

\begin{figure}
\includegraphics{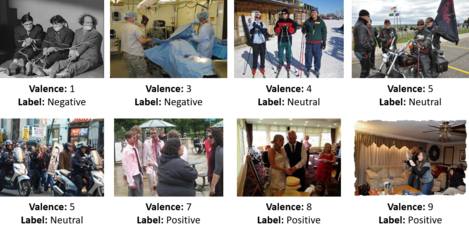}
\caption{Some Samples of the EmotiC Dataset}
\label{fig:emoticsamples}
\end{figure}

We evaluate 2 variations.  The first model (Global$\_$Simple) is obtained by initializing its weights with those of the pre-trained DenseNet-161 model and fine-tune on the EmotiW 2018 dataset. Similarly, the second model (Global$\_$EmotiC) is also initialized with the pre-trained DenseNet-161 weights. However, before being fine-tuned with the EmotiW 2018 dataset, images from the EmotiC dataset \cite{emotic_cvpr2017} are employed for the first training.  Some samples of this dataset are shown in Fig.~\ref{fig:emoticsamples}.  It contains 23,554 images in total.  The valence labels of this dataset are marked as negative, neutral or positive by demarcating its range of 1-10 into 1-3, 4-6 and 7-10 respectively. 

In both models, we rescale all the images to have a minimum side of 256, and randomly crop 224$\times$224 regions. A batch-size of 32 is used, and the learning rate starts from 0.001. While the learning rate was divided by 10 every 7 epochs when fine-tuning on EmotiW 2018, we reduced it by the same factor after 6 epochs when using EmotiC. 




The results of the Global-Level CNN on the EmotiW 2018 dataset are presented in Table \ref{tab:modelresults}. Pre-training the Global-Level CNN on the EmotiC dataset improves the performance of this model. This is primarily due to the fact that the EmotiC dataset contains images similar to EmotiW, but is larger and therefore helps to avoid over-fitting.



\subsubsection{Local-Level CNN}\label{sec:localcnneval}

We evaluate 3 variations of the local-level CNN. All three variations are pre-trained to recognize the emotion of a given face, and not of the image as a whole.  For the first and second models (Local and Local$\_$FineTune), we use a batch-size of 60, and the learning rate starts from 0.01, being divided by 10 every 7 epochs for 25 epochs.  For the last model (Local$\_$FineTune$\_$LSoftmax), we use L-softmax \cite{Liu:2016:LSL:3045390.3045445}.

After pre-training, Local predicts the emotion of the image by taking an average of the model output for each face in the image. Local$\_$FineTune and Local$\_$FineTune$\_$LSoftmax are fine-tuned to recognize the emotion of the image, given its cropped aligned faces.  A batch-size of 32 is used for fine-tuning, and the learning rate starts from 0.001, being divided by 10 every 4 epochs for 12 epochs.

The results of the Local-Level CNN on the EmotiW 2018 dataset are presented in Table \ref{tab:modelresults}.  Local$\_$FineTune achieves the best performance, which shows that fine-tuning the model to predict emotions of the images after training it for emotion recognition of faces gives a better generalization.  Further, the poor performance of Local$\_$FineTune$\_$LSoftmax can be attributed to the use of L-softmax loss, as it is highly unstable during training.



\begin{table}[]
\caption{Results on the Validation Set}
\label{tab:modelresults}
\small
\begin{tabular}{clcc}
\toprule
  & Model   & VAL & EVAL \\ \midrule
\multirow{2}{*}{\begin{tabular}[c]{@{}c@{}}Global-Level \\ CNN\end{tabular}} 
& Global$\_$Simple  &69.50\% & 70.80\%\\
& Global$\_$EmotiC  &70.40\% & 70.20\%\\ \midrule
\multirow{3}{*}{\begin{tabular}[c]{@{}c@{}}Local-Level \\ CNN\end{tabular}} 
& Local &71.18\% & 72.40\%  \\
& Local$\_$FineTune &71.51\% & 74.20\% \\
& Local$\_$FineTune$\_$LSoftmax &68.96\% & 69.90\% \\ \midrule
\multirow{6}{*}{\begin{tabular}[c]{@{}c@{}}Attention \\ Models\end{tabular}}  
& Average &73.03\% & 73.90\% \\
& Attention$\_$A &73.18\% & 73.00\% \\
& Attention$\_$B &73.75\% & 75.10\% \\
& Attention$\_$B$\_$EmotiC  &74.26\% & 75.20\% \\
& Attention$\_$C &\textbf{74.38}\% & 75.00\% \\
& Attention$\_$C$\_$EmotiC &73.66\% & \textbf{76.20}\% \\
\bottomrule
\end{tabular}
\end{table}

\subsubsection{Attention Models}


We evaluate 6 variations of the attention model. All the variations use pre-trained weights of Local as the local branch and either Global$\_$Simple or Global$\_$EmotiC as the global branch. In addition, dropout \cite{JMLR:v15:srivastava14a} is used after every fully connected layer (except the output layer), and batch normalization \cite{DBLP:journals/corr/IoffeS15} is employed separately on the global and weighted sum of face features just before concatenation.

The first model (Average) computes the average features with the same importance to each face. The second model (Attention$\_$A) uses an attention scheme based on a global image feature vector representation (see Sec.~ \ref{sec:mechanism2}).  The face feature vectors and the global image feature vector are 256-dimensional. The third (Attention$\_$B) and fourth (Attention$\_$B$\_$EmotiC) models use an intermediate transformation of the global image feature vector (see Sec.~\ref{sec:mechanism3}). The face feature vectors and the intermediate feature vector are 64-dimensional in this case. The fifth (Attention$\_$C) and sixth (Attention$\_$C$\_$EmotiC) models use the attention mechanism based on the generation of a score through a fully connected network (see Sec.~ \ref{sec:mechanism4}). The face feature vectors are 256-dimensional, which are passed through a simple fully connected neural network containing 2 layers with 64 nodes and 1 node respectively.

A batch size of 32 and learning rate of 0.001 is used for all the models. Learning rate decay is applied, the learning rate being divided by 10 every 5 epochs for 16 epochs in the first model, whereas in the rest of the models, it is divided by 10 every 9 epochs for 27 epochs.

The results of the different attention models on the EmotiW 2018 dataset are reported in Table \ref{tab:modelresults}.  Training an end-to-end attention model gives a 1-2\% rise in validation performance.  The general trend in performance is Attention C $>$ Attention B $>$ Attention A = Average.  Also, taking the Global$\_$EmotiC model rather than the Global$\_$Simple model as the global branch gives a better performance, 
as can be seen by the evaluation performance of Attention$\_$B$\_$EmotiC and Attention$\_$C$\_$EmotiC.

\subsubsection{Final Submissions}

In Table \ref{tab:submissionResults} we present results of our submissions on the test dataset of EmotiW 2018 challenge. 
We report on our best attention model, i.e. Attention$\_$C$\_$EmotiC (Single) as well as an ensemble of 14 models (Ensemble) obtained from the different configurations and hyper-parameters of models mentioned in the previous experiments. 
For the ensemble, the final classification decision is computed by averaging the probabilities of the employed models. All models submitted were trained on the training subset as well as on the VAL subset.  We notice a significant gap in the validation and submission test accuracy.  This is most probably due to a domain shift and it requires further investigation. 

\begin{table}
  \caption{Submission Results}
  \label{tab:submissionResults}
  \begin{tabular}{cccccc}
  \toprule
Model & EVAL \footnotemark & \multicolumn{4}{c}{EmotiW 2018 Test Dataset} \\
    \midrule
    \ & & Positive & Neutral & Negative & Overall \\
    \ Single & 78.20\% &66.59\% & 57.97\% &58.87\% & 61.84\%\\
    \ Ensemble & 80.90\% &71.33\% & 60.48\% &59.71\% & \textbf{64.83\%}\\

    \bottomrule
  \end{tabular}
\end{table}

\footnotetext[1]{Note that here the EVAL performance is higher because during training, in Table 1, we use training data, whereas in Table 2, we use training data + VAL data}
\addtocounter{footnote}{2}

\section{Conclusions}
We have presented a method to improve the classification performance of group-level emotions in images, which we use as our approach for the EmotiW 2018 group-level emotion recognition challenge.  The main contribution of the paper lies in the use of an attention mechanism to merge the features representing the different faces present in an image. This local representation is then fused with a global representation in an end-to-end trainable model.  We also explored using a larger similar dataset and combinations of different variations of these models. 

For future work, we plan to explore other cues of the image for a more robust emotion recognition. For instance, the pose and context of the people can help to better understand what is happening in the image and therefore to better estimate the group-level emotion. 


\bibliographystyle{ACM-Reference-Format}
\bibliography{emotiW_bibliography}-

\end{document}